\documentclass{article}

\usepackage{PRIMEarxiv}

\usepackage[utf8]{inputenc} 
\usepackage[T1]{fontenc}    
\usepackage{hyperref}       
\usepackage{url}            
\usepackage{booktabs}       
\usepackage{amsfonts}       
\usepackage{nicefrac}       
\usepackage{microtype}      
\usepackage{lipsum}
\usepackage{fancyhdr}       
\usepackage{graphicx}       
\graphicspath{{media/}}     

\usepackage{amsmath,amssymb,bm}
\usepackage{subcaption}
\usepackage{tabularx}

\title{TriHorn-Net: A Model for Accurate Depth-Based 3D Hand Pose Estimation}

\author{
  Mohammad Rezaei \\
  University of Texas at Arlington\\
  \texttt{mohammad.rezaei@mavs.uta.edu} \\
   \And
  Razieh Rastgoo* \\
  Semnan University \\
  \texttt{rrastgoo@semnan.ac.ir} \\
  \texttt{*Corresponding Author}\\
   \And
  Vassilis Athitsos \\
  University of Texas at Arlington\\
  \texttt{athitsos@uta.edu} \\ 
}

\begin{document}
\maketitle

\begin{abstract}
3D hand pose estimation methods have made significant progress recently. However, estimation accuracy is often far from sufficient for specific real-world applications, and thus there is significant room for improvement. This paper proposes TriHorn-Net, a novel model that uses specific innovations to improve hand pose estimation accuracy on depth images. The first innovation is decomposition of the 3D hand pose estimation into the estimation of 2D joint locations in the depth image space (UV), and the estimation of their corresponding depths aided by two complementary attention maps. This decomposition prevents depth estimation, which is a more difficult task, from interfering with the UV estimations at both the prediction and feature levels. The second innovation is PixDropout, which is, to the best of our knowledge, the first appearance-based data augmentation method for hand depth images. Experimental results demonstrate that the proposed model outperforms the state-of-the-art methods on three public benchmark datasets. Our implementation is available at \url{https://github.com/mrezaei92/TriHorn-Net}.
\keywords{3D Hand Pose Estimation, Depth Image, Accuracy, 2D joints, Attention.}
\end{abstract}

\section{Introduction}
Hands are crucial in allowing humans to interact with the world around them. Accurate hand pose estimation has many applications in areas such as human computer interaction (HCI), augmented reality (AR), virtual reality (VR) and gesture recognition \cite{b1,b2,b3}. As commodity depth cameras become more accurate and affordable and, as a result, more widely used, significant advancements have been made in depth-based 3D hand pose estimation \cite{yuan2018depth,malik2020handvoxnet,chen2019so,du2019crossinfonet,xiong2019a2j,ge2018point,wan2018dense,baek2018augmented} and hand segmentation \cite{tompson2014real,bojja2019handseg}. However, 3D hand pose estimation remains a very challenging task due to the large degree of variation in hand appearance, heavy self-occlusion, noise,  high dimensionality and self-similarity between hand parts \cite{du2019crossinfonet,ge2018hand,ge2018point,moon2018v2v}.
\par
With the advancement of deep neural networks (DNNs) \cite{b7}, DNN-based hand pose estimation techniques rapidly displaced the previous methods such as \cite{keskin2012hand,oikonomidis2011efficient} and have achieved impressive results. These methods can be broadly categorized into two groups: 1)  regression-based methods and 2) detection-based methods. Regression-based methods \cite{ge2018hand,oberweger2015hands,poier2019murauer} encode the hand depth image into a single global feature which is in turn used to directly estimate the hand joints. Detection-based methods adopt a dense-prediction approach, where they utilize hierarchical features to compute pixel-wise predictions for each joint. Detection-based methods generally tend to outperform regression-based methods \cite{yuan2018depth} because regression-based methods use a single global feature for estimation, which cannot fully retain fine-grained spatial information required for accurate mapping into 3D hand poses. 
\par
Despite the superior performance of detection-based methods, they still suffer from several drawbacks. Moon et al.~\cite{moon2018v2v} achieve a high accuracy by using 3D CNNs, but their method comes at a heavy computational and memory cost. A new class of detection-based methods has recently emerged that adopts an approach based on dense pixel or point offset prediction, whereby they densely estimates all the pixels’(or points') offsets to joints and compute joint positions by a weighted average over all the corresponding offset values. Despite their high performance, \cite{li2019point,ren2019srn} use non-learnable information aggregation operations such as argmax operation or mean-shift estimation to compute joint coordinates from the heatmap or offset vector fields. However, the information  aggregation operation is treated as a post-processing step and is not incorporated into the training phase, causing a gap between training and inference. \cite{wan2018dense,ge2018hand} require complex post-processing operations, such as taking neighboring points, causing inevitable quantization errors and rendering the pipeline not end-to-end differentiable. JGR-P2O~\cite{fang2020jgr} partly solves these issues by predicting pixel-wise offsets and a weight map to compute the joint positions using the weighted average, but it still suffers from the common issue among this class of methods that the estimations are unstable when depth values near the target joint are heavily missing, leading to a performance degradation.
\par
To tackle the aforementioned issues, we propose a novel model, that we call TriHorn-Net, for 3D hand pose estimation. TriHorn-Net consists of an encoder network that encodes the input hand depth image into a high-resolution feature volume, and three separate branches that take the hand feature volume as the input and together estimate the 3D hand pose. The first two branches compute two per-joint attention maps that are fused subsequently. The two attention maps are complementary in the sense that one guides the network's attention towards the pixels where the joints occur, and the other guides the network's attention towards non-joint pixels that can potentially give the network useful clues for estimating the corresponding joint depths. The attention maps computed by the UV branch are explicitly encouraged to focus on joint pixels by applying 2D supervision to the heatmaps resulted from passing them through a spatial softmax layer \cite{iqbal2018hand}. This approach can be viewed as the typical detection-based approach based on dense pixel-wise joint predictions. The attention maps computed by the attention enhancement branch are learned under no constraints, allowing them to freely focus on hand pixels most relevant to the estimation of joint depths. The depth branch develops pixel-wise feature vectors that contain depth information of the joints. The proposed model uses the fused per-joint attention maps as guidance to pool features from relevant pixels for each joint. After the relevant features are pooled for each joint, a weight-sharing linear layer is used to estimate the corresponding depth value.
\par
We also propose PixDropout, a simple yet effective appearance-based data augmentation function for depth-based hand pose estimation methods. This function performs augmentation on a given sample by uniformly sampling a fraction of the pixels on the hand surface and turning them into a background pixel (replaces their value with a constant background value). We show empirically that PixDropout leads to a performance improvement not only in the proposed method but also in a regression-based method.  
\par

The proposed model is end-to-end differentiable and does not include any post-processing step or data pre-processing such as converting the depth map into point clouds \cite{chen2019so,ge2018hand,cheng2021handfoldingnet} or voxelized volume\cite{moon2018v2v}.
We conduct the evaluation of the proposed model on three publicly available datasets, namely ICVL \cite{tang2014latent}, MSRA \cite{sun2015cascaded} and NYU \cite{tompson2014real}, which are challenging benchmarks commonly used for evaluation of 3D hand pose estimation methods. The results demonstrate that the proposed model outperforms the state-of-the-art methods on all these benchmarks.

In summary, our contributions are as follows:
\begin{itemize}
  \item We propose a novel neural network architecture, TriHorn-Net, which enables accurate 3D hand pose estimation.
  \item We propose a novel formulation for effective decomposition of the hand pose estimation into the estimation of the 2D joint locations and their depths. 
  \item We propose PixDropout, which is, to the best of our knowledge, the first appearance-based data augmentation function for depth-based hand pose estimation methods.
  \item We conduct extensive experiments to demonstrate that the proposed method outperforms the state-of-the-art methods. Our implementation is available at \url{https://github.com/mrezaei92/TriHorn-Net}.
\end{itemize}

\section{Related Work}
\subsection{3D Hand Pose Estimation}
Hand pose estimation has been a long-standing problem in Computer Vision. Before the widespread use of deep learning techniques, many approaches relied on hand-crafted features, optimization methods, and distance metrics. Athitsos et al. \cite{athitsos2003estimating} used edge maps and Chamfer matching to perform 3D hand pose estimation. Other approaches used optimization methods such as Particle Swarm Optimization (PSO) \cite{sharp2015accurate,oikonomidis2011efficient}. After the rise of deep learning, DNN-based methods quickly displaced the traditional methods. The two most common input data modalities for DNN-based methods are: 1) RGB images and 2) depth images. While DNN-based 3D hand pose estimation on RGB images is a relatively new field of research, it has attracted a lot of attention recently \cite{zimmermann2017learning,tang2021towards,zhou2020monocular,ge20193d,baek2019pushing,zhang2021hand,moon2020deephandmesh,moon2020i2l,spurr2020weakly}. However, as our method is depth-based, we focus our attention here on other depth-based methods.
\par
Depth-based hand pose estimation methods have significantly advanced in the last decade. These methods can be classified into three categories: generative methods \cite{khamis2015learning,romero2017embodied,tkach2017online,tzionas2016capturing}, discriminative methods \cite{keskin2012hand,liang2014parsing}, and hybrid methods \cite{sharp2015accurate,tang2015opening,taylor2016efficient}. Oberweger et al.~\cite{oberweger2015hands} used a CNN to estimate the hand pose represented by PCA coefficients. Instead of performing in the 2.5D space, several methods~\cite{ge20173d,moon2018v2v} converted 2.5D depth images into 3D voxels and adopted 3D CNNs to estimate the 3D hand pose. Fang et al.~\cite{fang2020jgr} propose an approach based on graph CNNs to compute pixel offsets to the joints and use weighted average to compute the hand joint locations. Another line of research has recently emerged, that utilizes the latest advancements of point cloud processing, by converting depth images into point clouds and using a point cloud processing network to perform hand pose estimation \cite{ge2018hand,ge2018point,li2019point,chen2019so,cheng2021handfoldingnet,huang2020hand}.
\par
TriHorn-Net is inspired by the methods based on dense pixel-wise prediction, but it differs from them in some important aspects. It offers a novel formulation for hand pose estimation, which is based on the decomposition of hand pose estimation into estimating 2D joint locations and their depth values. While it takes advantage of the typical pixel-wise prediction approach for estimating the 2D joint locations, it breaks from the standard approach in the sense that it estimates pixel-wise feature vectors (as opposed to predictions) and uses a weight-sharing layer (as opposed to dedicated layers) for estimating the joint's depth values. Extensive experiments demonstrate the effectiveness of the proposed formulation.
\par
The proposed model is similar to A2J~\cite{xiong2019a2j} in that it uses different branches for estimating joints' image coordinates and their depth values. However, it adopts a fundamentally different approach for estimating the joint positions. A2J~\cite{xiong2019a2j} relies on a fixed number of regularly spaced points placed in depth image space, which are called anchors, in order to predict joint UVD offsets, whereas the proposed method performs a pixel-wise likelihood estimation for computing joints' image coordinates UV, and uses the UV estimations to guide the estimation of the joint depth values. 
\par
 
\subsection{Data Augmentation}
Data augmentation methods aim at increasing the amount and diversity of the training data by randomly creating novel and realistic-looking data samples. In recent years, significant progress has been made on data augmentation methods for vision~\cite{krizhevsky2012imagenet,cubuk2018autoaugment}, NLP~\cite{yu2018qanet} and speech~\cite{hannun2014deep,park2019specaugment}. In the image domain, novel samples are created by applying a set of transformations to an available sample. These transformations can be broadly categorized into two groups: 1) geometric transformations and 2) appearance transformations. There has been a wide range of appearance transformations proposed recently for performing data augmentation on RGB images, such as color jitter, histogram equalization and contrast adjustment. However, most of them are not applicable to depth images due to the different nature of the depth image, limiting the set of data augmentation functions used by depth-based hand pose estimation methods mostly to geometric transformations such as random rotation, translation and scaling.
\par
In this paper, we propose PixDropout, a simple yet effective appearance transformation applicable as a data augmentation function to the depth images. It is strongly inspired by Dropout~\cite{srivastava2014dropout}. While Dropout~\cite{srivastava2014dropout} randomly selects and drops neurons in the layers of a neural network, the proposed PixDropout randomly drops some fraction of pixels on the hand surface in the input depth image. PixDropout is most similar to the RGB image augmentation methods proposed in \cite{devries2017improved,zhong2020random}, where they randomly sample and then mask out rectangular regions of the input image to simulate occlusion for an image recognition task. However, in contrast to \cite{devries2017improved,zhong2020random}, PixDropout applies no spatial priors(e.g., it samples individual pixels rather than rectangular regions). Empirical results show that despite its simplicity, PixDropout is an effective data augmentation function for depth-based hand pose estimation methods.

\begin{figure}[t]
\centering
\includegraphics[height=5cm]{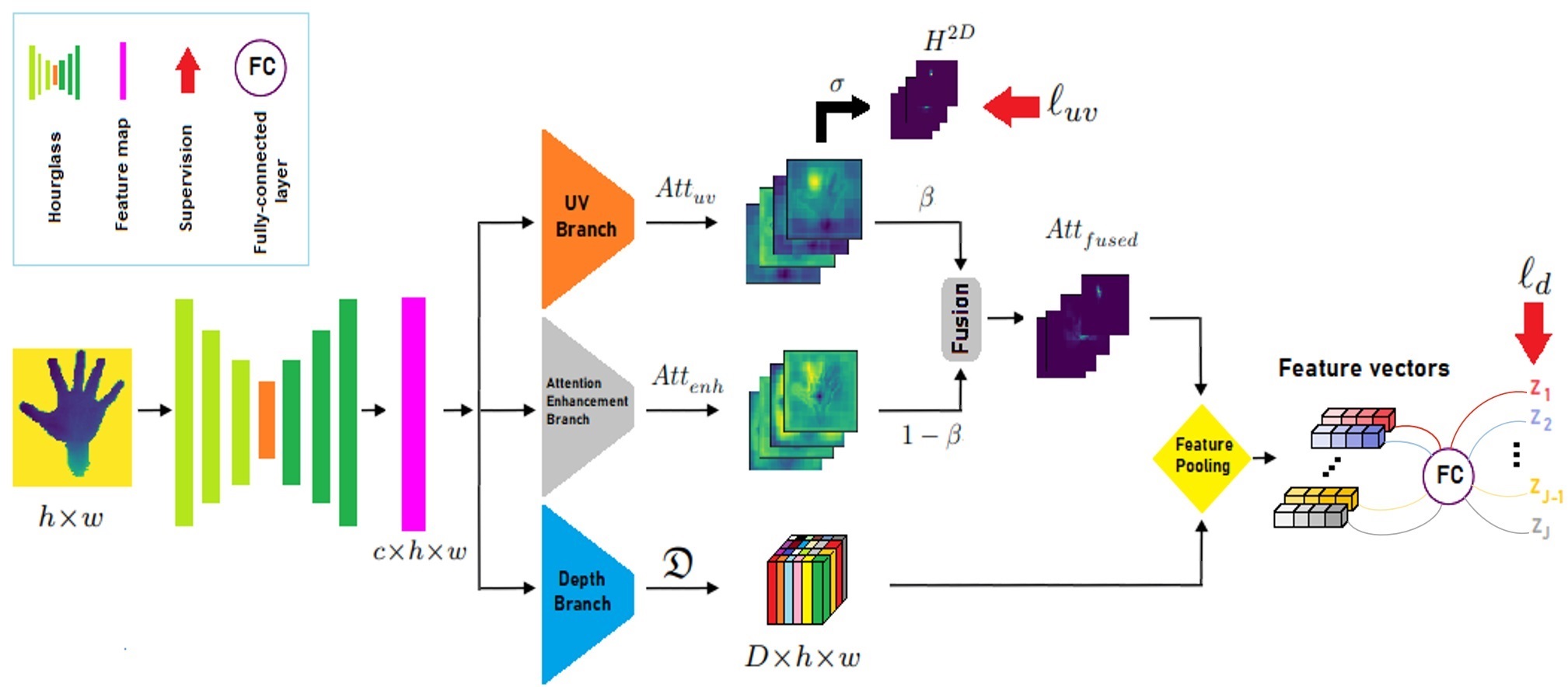}
\caption{An overview of the TriHorn-Net architecture. It consists of an encoder that encodes the hand depth image into a high resolution feature volume, which serves as the input to three separate branches. The UV branch and the attention enhancement branch compute two per-joint attention maps $Att_{uv}$ and $Att_{enh}$ respectively. $Att_{uv}$ is focused on joint pixels, whereas $Att_{enh}$ attention map has the flexibility to shift the network's attention to hand pixels that are most relevant for joint depth estimation. $Att_{uv}$ and $Att_{enh}$ are fused via a linear interpolation controlled by per-joint learned parameters $\beta_j$. For estimating the joints' depths, the network uses the fused attention maps $Att_{fused}$ to pool features from the depth feature map $\mathfrak{D}$ computed by the depth branch. The pooled feature vector for each joint is input to a weight-sharing linear layer to estimate its depth value.}
\label{fig:overview_main}
\end{figure}

\section{Methodology}
The task of 3D hand pose estimation is defined as follows: given an input depth image $D_I\in{\mathbb{R}^{H\times{W}}}$, the task is to estimate the 3D location of a set of pre-defined hand joints $P\in{\mathbb{R}^{J\times{3}}}$ in the camera coordinate system. $H$, $W$ denote the height and width of the input depth image respectively. $J$ denotes the total number of joints to be estimated. In this section, the proposed model is laid out in details.
\par
As illustrated in Figure \ref{fig:overview_main}, TriHorn-Net consists of two stages. In the first stage, the input depth image is run through the encoder network $f$. The encoder extracts and combines low-level and high-level features of the hand and outputs a high resolution feature volume, which is passed on to three separate branches. The UV branch, computes a per-joint attention map, where each map is focused on pixels where the corresponding joint occurs. This behavior is explicitly enforced by the application of 2D supervision to the heatmaps computed by passing the attention maps through a special softmax layer \cite{iqbal2018hand}. The second branch, called the attention enhancement branch, also computes a per-joint attention map but does so under no constraints, allowing it to freely learn to detect the hand pixels most important for estimating the joint depth values under different scenarios. This attention map enhances the attention map computed by the UV branch through a fusion operation, which is performed by a linear interpolation controlled by per-joint learnable parameters. As a result, the fused attention maps attend to not only the joint pixels but also the hand pixels that do not belong to joints but contain useful information for estimating the joint depth values. The fused attention map is then used as guidance for pooling features from the depth feature map computed by the depth branch. Finally, a weight-sharing linear layer is used to estimate the joint depth values from the feature vectors computed for each joint.
\par
The feature pooling is conducted through a dot-product operation. This type of feature pooling followed by an estimation layer shared across all the joints is adopted from Transformer networks \cite{vaswani2017attention}. This approach breaks with the standard approach in hand pose estimation methods, where they use dedicated layers for estimating the location of each joint. We show in the sec~\ref{subsection:ablation} that TriHorn-Net derives its power from the proposed estimation formulation, namely the decomposition of the hand pose estimation into the estimation of 2D joint locations and the estimation of their corresponding depth values aided by two separate attention maps.

\subsection{Encoder} \label{subsection:encoder}
The encoder is defined as a non-linear mapping from the input depth image to the output feature volume $f: {\mathbb{R}^{H\times{W}}}\rightarrow {\mathbb{R}^{c\times{h\times{w}}}}$, where $h$, $w$ and $c$ denote the height, width and the number of the channels of the output feature volume respectively. While any off-the-shelf network architecture can be used as this non-linear mapping, we empirically find that, to maximize accuracy, the encoder network should have a high capability of extracting and fusing features at different scales. This is because the hand orientation, the arrangement of the fingers, and the relationships of adjacent joints are among the many cues that are best recognized at different scales in the depth image. We show in the sec~\ref{subsection:ablation} that the proposed model is robust to the choice of the encoder network architecture as long as the above-mentioned requirement of extracting and fusing features at different scales is met. We use an Hourglass network~\cite{newell2016stacked} as the encoder. It uses skip-connections and repeated bottom-up, top-down processing to extract and consolidate features across different scales. 

\subsection{UV Branch}
This branch takes as input the output feature volume from the encoder and computes a per-joint attention map $Att_{uv} \in{ \mathbb{R}^{J\times{h\times{w}}} }$. We use $Att_{uv}^j \in{ \mathbb{R}^{h\times{w}} }$ to refer to the attention map corresponding to the $j^{th}$ joint. $Att_{uv}^j$ is explicitly encouraged to focus on pixels where the $j^{th}$ joint occurs by applying 2D supervision to it. To this end, the attention map $Att_{uv}^j$ is first normalized by a spatial softmax layer \cite{iqbal2018hand} to obtain the corresponding heatmap ${H}_j^{2D} = \sigma(Att_{uv}^j)$ as follows:

\begin{equation}
  {H}_j^{2D}(x,y) =\frac{exp(Att_{uv}^j(x,y))}{\sum\limits_{u_i,v_i\in{\Omega}}exp(Att_{uv}^j(u_i,v_i))}
\end{equation}

In the above, $\sigma$ denotes the spatial softmax layer. The heatmap $H_j^{2D}$ represents the likelihood of the $j^{th}$ joint occurring at each pixel location. $\Omega$ represents the spacial domain of the attention map $Att_{uv}^j$. The 2D location of the $j^{th}$ joint is computed through an integration operation similar to \cite{sun2018integral,iqbal2018hand}, as follows: 

\begin{equation}
  (\bar{U}^j,\bar{V}^j) = \sum_{u_i}\sum_{v_i}(u_i,v_i){H}_j^{2D}(u_i,v_i)
\end{equation}

In the above, $(\bar{U}^j,\bar{V}^j)$ represents the estimated coordinates of the $j^{th}$ joint in the depth image space. The supervision is applied to all attention maps $Att_{uv}^j$ for $j \in{1,2,...,J}$ by minimizing the mean L1 distance defined as:
\begin{equation}
  \ell_{uv} = \frac{1}{2J}\sum_{j=1}^J|\bar{U}^j-U^j|+|\bar{V}^j-V^j|
\end{equation}
In the above, $(U^j,V^j)$ represents the ground-truth 2D location of the $j^{th}$ joint. Note that $(\bar{U}^j,\bar{V}^j)$ is also used to report the estimated coordinates of the $j^{th}$ joint in the depth image space. 

\subsection{Attention Enhancement Branch}
This branch is aimed at computing a more flexible attention map to enhance the attention maps computed by the UV branch $Att_{uv}$ towards facilitating the estimation of the joint depth values. Specifically, it takes as input the output feature volume from the encoder and computes a per-joint attention map $Att_{enh} \in{ \mathbb{R}^{J\times{h\times{w}}} }$. $Att_{enh}^j \in{ \mathbb{R}^{h\times{w}} }$ denotes the attention map corresponding to the $j^{th}$ joint. In contrast to $Att_{uv}$ , no external constraint (supervision) is applied to this attention map, allowing it to freely learn which hand pixels are the most relevant ones for estimating the depth value for each joint under different scenarios.

\subsection{Depth Branch}
Contrary to the common practice of computing pixel-wise depth offset or prediction, the proposed model computes dense pixel-wise depth feature vectors. Specifically, this branch takes as input the output feature volume from the encoder and produces a dense depth feature map $\mathfrak{D} \in{\mathbb{R}^{D\times{h\times{w}}}}$. $D$ represents the depth feature vector dimension, which is set $D = 64$ in our experiments. The depth feature vector at the special location $(x,y)$ in the depth feature map $\mathfrak{D}$, denoted by $\mathfrak{D}(x,y) \in{\mathbb{R}^D}$, is developed such that it contains information about the depth value of the joints gathered from the input depth image pixels included in the receptive field of the special location $(x,y)$ in the depth feature volume. The final feature vector used for each joint to estimate its depth value is obtained using a weighted average computed over all the depth feature vectors, where the weight to each depth feature vector is assigned using the corresponding fused attention map.


\subsection{Attention Fusion}
\begin{figure}[t]
\centering
\includegraphics[height=4cm]{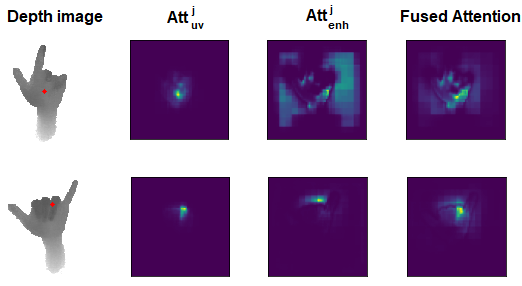}
\caption{Qualitative examples of how the two attention maps complement each other to guide feature pooling for depth estimation. Each row shows an example including the input depth image, the $Att_{uv}^j$ and $Att_{enh}^j$ attention maps computed by the first and second branches respectively, and the resulting fused attention map $Att_{fused}^j$. The red dot in the input depth image marks the ground-truth joint location for which the attention maps are computed}
\label{fig:attention}
\end{figure}

The two attention maps $Att_{uv}^j$ and $Att_{enh}^j$ are complementary in the sense that $Att_{uv}^j$ shifts the network attention to the pixels where the $j^{th}$ joint occurs, while $Att_{enh}^j$ helps the network pay attention to the non-joint pixels that might contain useful information for estimating the depth value of the $j^{th}$ joint. These two attention maps are fused as follows:
\begin{equation}
  Att_{fused}^j = \sigma(\beta_j Att_{uv}^j + (1 - \beta_j)Att_{enh}^j)
\end{equation}
Here, $\beta_j \in{[0,1]}$ denotes the learned parameter that controls the contribution of each attention map to the fused attention map $Att_{fused}^j$. The proposed model uses $Att_{fused}^j$ as guidance to pool features from the pixels that contain the most relevant information with respect to the depth of the $j^{th}$ joint. Figure~\ref{fig:attention} shows some qualitative examples of how the two attention maps $Att_{uv}^j$ and $Att_{enh}^j$ play the complementary role in forming the final fused attention map $Att_{fused}^j$ in order to guide the subsequent feature pooling for depth value estimation.

\subsection{Depth Value Estimation}
The depth value for the $j^{th}$ joint is estimated from the feature vector obtained by pooling features from the pixels that contain the most relevant information about its depth, which is guided by $Att_{fused}^j$ as follows:
\begin{equation}
  F_j = Att_{fused}^j \circ{\mathfrak{D}} =  \sum_{x}\sum_{y} Att_{fused}^j(x,y) \mathfrak{D}(x,y)
\end{equation}
Where $F_j \in{\mathbb{R}^D}$ denotes the pooled feature vector for the $j^{th}$ joint. The depth value for the $j^{th}$ joint, denoted by $\bar{Z}_j$, is then estimated using a single linear layer as follows:
\begin{equation}
  \bar{Z}_j = F_j \bm{W} + \bm{b}
\end{equation}
Where $\bm{W} \in{\mathbb{R}^{D}}$ and $\bm{b} \in{\mathbb{R}}$ denote the weights of the linear layer. Note that this linear layer is shared across all the joints. This not only improves the parameter efficiency but also encourages ensemble-like behavior in the depth feature vectors, as each has to gather the depth information of all joints. This type of feature pooling using attention followed by a shared layer is inspired from the mechanism employed in Transformer networks~\cite{vaswani2017attention}.
\par
The depth value estimation for the joints is supervised by the following loss term:
\begin{equation}
  \ell_{d} = \frac{1}{J}\sum_{j=1}^J|\bar{Z}_j - Z_j|
\end{equation}
Where $Z_j$ refers to the ground-truth depth value for the $j^{th}$ joint. 

\subsection{End-to-End Training}
The proposed model is end-to-end differentiable and is trained by minimizing the loss function that comprises the two loss terms discussed in the previous sections, which is formulated as:
\begin{equation}
  \mathcal{L} = \ell_{uv} + \lambda \ell_{d}
\end{equation}
where $\lambda$ is a weighting factor to balance $\ell_{uv}$ and $\ell_{d}$. We set  $\lambda = 1$ in our experiments.

\subsection{PixDropout}
The proposed data augmentation function PixDropout is defined as a function $\mathcal{T}_{\alpha}: {\mathbb{R}^{H\times{W}}}\rightarrow {\mathbb{R}^{H\times{W}}}$, where the parameter $\alpha\in{[0,1]}$ controls the intensity of the augmentation. In the first step, we uniformly sample a probability $\gamma$ from the range $[0,\alpha]$. In the second step, we uniformly sample a set of the pixels $\mathcal{Q}$ from the hand surface. Each pixel is selected with a probability of $\gamma$. The augmented depth image $\hat{D_I} = \mathcal{T}_{\alpha}(D_I)$ is computed by dropping the selected pixels as follows: 
\begin{equation}
  \hat{D_I}(p)=
 \begin{cases}
    C & \text{if }  p\in{\mathcal{Q}}\\
    D_I(p)               & \text{otherwise}
\end{cases}
\end{equation}
Where $p$ denotes an arbitrary pixel in the depth image. $C$  represents the constant value assigned to the background (non-hand) pixels. 

\section{Experiments}
\subsection{Implementation Details}
The pre-processing method for preparing the input depth image includes first cropping the hand area from a depth image similar to \cite{oberweger2017deepprior++}, and then resizing it to a fixed size of 128x128. The depth values are normalized to [-1, 1]. In order to maximize accuracy, the encoder's output feature volume needs to be of high spatial dimension. To strike a balance between the computational complexity and performance, we set it to be half of that of the input depth image. We use Adam \cite{kingma2014adam} optimizer with a cosine learning rate decay schedule \cite{loshchilov2016sgdr} for training. The initial learning rate and the weight decay are set to be $10^{-3}$ and $10^{-5}$ respectively.  For data augmentation, we use geometric transformations including in-plane rotation ([-180, 180] degree), 3D scaling ([0.9, 1,1]), and 3D translation ([-8, 8] mm), as well as the proposed PixDropout as the appearance transformation. We set $\alpha = 0.15$ in all the experiments. We trained the model for 40 epochs on ICVL, 40 epochs on NYU and 60 epochs on MSRA. All experiments are implemented by PyTorch framework \cite{paszke2019pytorch} and conducted on a single server with one NVIDIA 1080Ti GPU.

\begin{table}
\caption{The performance of the models with and without (w/o) PixDropout as augmentation function on ICVL~\cite{tang2014latent}. The numbers indicate the mean distance error (mm)}
\centering
\setlength{\tabcolsep}{10pt}
\renewcommand{\arraystretch}{1.05}
\begin{tabular}{ccc}
            \hline
            Model & with PixDropout    &   w/o PixDropout       \\
            \hline
                TriHorn-Net          &         \textbf{5.73}     &  5.90    \\
                ResNet-50            &         \textbf{7.71}     &  7.81     \\

            \hline
        \end{tabular}
\label{tab:PixDropout}
\end{table}
\begin{table}[t]
\caption{Impact of using different attention maps for depth feature pooling on the performance on ICVL~\cite{tang2014latent}}
\centering
\setlength{\tabcolsep}{10pt}
\renewcommand{\arraystretch}{1.05}
\begin{tabular}{cc}
            \hline
            Attention Map    &   Error (mm)       \\
            \hline
                $Att_{uv}$           &         5.91   \\
                $Att_{enh}$            &         6.03  \\
                Fused Attention            &        \textbf{5.73}  \\

            \hline
        \end{tabular}
\label{tab:attention}
\end{table}
  
\subsection{Datasets and Evaluation Metrics}
\textbf{ICVL Dataset.} The ICVL dataset \cite{tang2014latent} provides 22K and 1.6K depth frames for training and testing, respectively. The ground-truth for each frame contains J = 16 joints, including one joint for the palm and three joints for each finger. We do not use the additional 300k augmented frames (which are obtained with in-plane rotations of the original training frames) included in this dataset.
\par
\textbf{MSRA Dataset.} The MSRA dataset \cite{sun2015cascaded} contains more than 76K frames captured from 9 subjects. Each subject contains 17 hand gestures and each hand gesture has about 500 frames with segmented hand depth image. Each frame is provided with a ground-truth of J = 21 joints, including one joint for the wrist and four joints for each finger. Following the protocol used by \cite{sun2015cascaded}, we evaluate the proposed method on this dataset with the leave-one-subject-out cross-validation strategy.
\par
\textbf{NYU Dataset.} The NYU dataset \cite{tompson2014real} is captured from three different views with Microsoft Kinect sensor. Each view contains 72K training 8K testing depth images. Following the common protocol, we only use the first view with a subset of J = 14 joints out of total of 36 annotated joints provided for both the training and testing.
\par
\textbf{Evaluation metrics.}
 We use the two most commonly used metrics for evaluation of 3D hand pose estimation: the mean distance error (in mm) and the success rate. The mean distance error measures the average Euclidean distance between the estimated and the ground-truth coordinates computed across all the joints and over the entire testing set. The success rate is defined as the fraction of the frames for which the mean distance error is less than a certain distance threshold.

 \begin{table}[t]
\caption{Comparison of different choices for the encoder network architecture on ICVL~\cite{tang2014latent}. \#Params indicates the number of the model parameters}
\centering
\setlength{\tabcolsep}{10pt}
\renewcommand{\arraystretch}{1.05}
\begin{tabular}{ccc}
            \hline
            Encoder Architecture  &   Error (mm)          & \#Params              \\
            \hline
                HRnet~\cite{wang2020deep}              &         5.91     &   7.22M        \\
                ResDeconv~\cite{xiao2018simple}        &         6.04     &   27.24M      \\
                Hourglass~\cite{newell2016stacked}     &  \textbf{5.73}    &   7.81M       \\
        
            \hline
        \end{tabular}
\label{tab:encoderArcitecture}
\end{table}

\begin{table}[t]
\caption{Comparison of different attention fusion approaches on ICVL~\cite{tang2014latent}}
\centering
\setlength{\tabcolsep}{10pt}
\renewcommand{\arraystretch}{1.05}
\begin{tabular}{ccc}
            \hline
            Approach  &   Error (mm)      \\
            \hline
                Concatenation          &         5.98    \\
                Summation      &         5.84    \\
                Ours     &  \textbf{5.73}   \\
        
            \hline
        \end{tabular}
\label{tab:attentionFusion}
\end{table}
 
\subsection{Ablation Study} \label{subsection:ablation}
\textbf{Impact of using complementary attention maps.} We study the impact of employing two complementary attention maps in the proposed model. Specifically, we examine the performance of the model in three cases with respect to the attention map used for depth feature pooling. The first case only uses the attention map computed by the UV branch and removes the attention enhancement branch from the network. The second case only uses $Att_{enh}$, which is computed by the attention enhancement branch. The third case corresponds to the proposed approach based on fusing the two complementary attention maps. As can be seen in Table~\ref{tab:attention}, using a second freely learned attention map enhancing the attention map computed by the UV branch leads to the best performing case.
           
\par
\textbf{Effectiveness of Different Approaches for Attention Fusion.} We study the effectiveness of the proposed attention fusion approach in our model. We implement the proposed model using three different approaches for attention fusion: 1) concatenation 2) summation and 3) the proposed strategy. For concatenation, the two attention maps are first concatenated and then passed through a number of convolutional layers to obtain the fused attention map. For the second experiment, the two attention maps are simply fused by the element-wise addition. The proposed strategy is an extension of the summation approach, where the contribution of each attention map to the fused attention map is controlled by a learned parameter. As can be seen in Table~\ref{tab:attentionFusion}, the proposed strategy performs best.
\par
\textbf{Impact of Different Encoder Network Architectures.} We analyse the impact of different encoder network architectures on the model performance. Specifically, we use three representative architectures: 1) Hourglass network~\cite{newell2016stacked}, 2) HRNet~\cite{wang2020deep} and 3) ResNetDeconv~\cite{xiao2018simple}. Although through different mechanisms, Hourglass network and HRNet both directly transfer the low-level features extracted in the early layers to deeper layers via direct skip-connections. On the other hand, ResNetDeconv down-samples the input to a low resolution feature map and then up-samples it back by a deconvolution head. As can be seen in Table~\ref{tab:encoderArcitecture}, the Hourglass network and HRNet both achieve the state-of-the-art results. Although ResNetDeconv does not have the same ability to extract and fuse features at different scales, it still performs strongly compared to the existing methods. These observations demonstrate that the proposed model is robust to the choice of the encoder network architecture and derives its superior performance from our novel formulation of pose estimation.
\subsection{Effectiveness of PixDropout}
To verify the effectiveness of the proposed data augmentation function PixDropout, we conduct two independent comparisons. We compare the performance of the proposed model with and without PixDropout. We also repeat this comparison using a regression-based method. Specifically, we use a ResNet-50 network~\cite{he2016identity}, with its last fully connected layer replaced by 2 fully connected layers to estimate the hand pose. As can be seen in Table~\ref{tab:PixDropout}, PixDropout leads to a performance improvement not only in the proposed model but also in a model of different nature, demonstrating its effectiveness as an augmentation function for depth-based hand pose estimation methods.

\begin{figure}
    \centering 
    \begin{subfigure}[b]{0.32\linewidth}  
        \centering 
        \includegraphics[width=\textwidth]{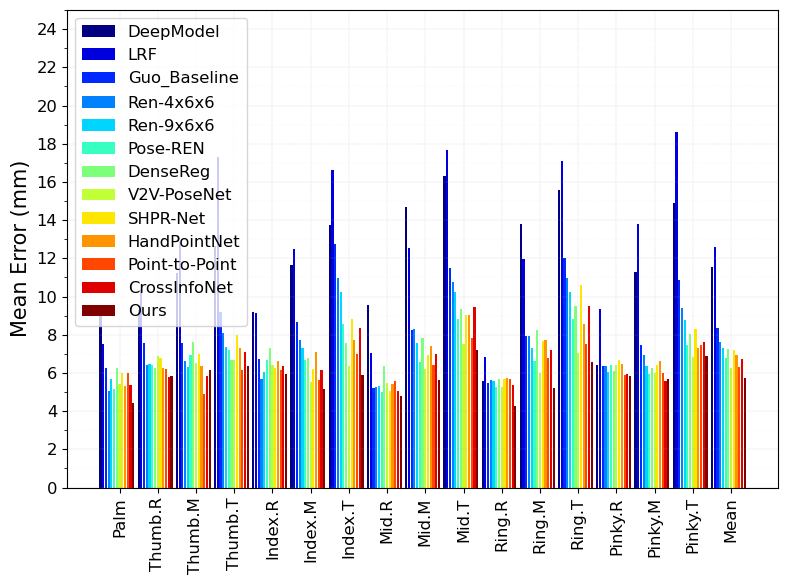} 
    \end{subfigure} 
    \begin{subfigure}[b]{0.32\linewidth} 
        \centering 
        \includegraphics[width=\textwidth]{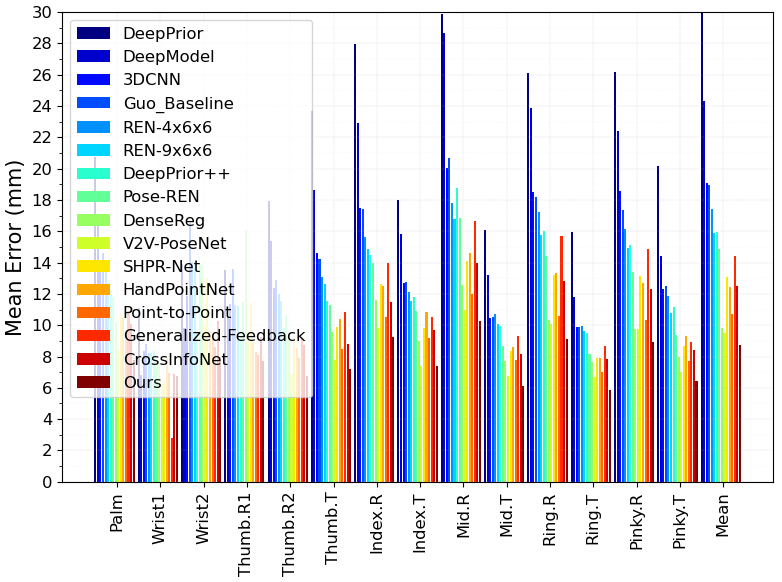} 
    \end{subfigure}
    \begin{subfigure}[b]{0.32\linewidth} 
        \centering 
        \includegraphics[width=\textwidth]{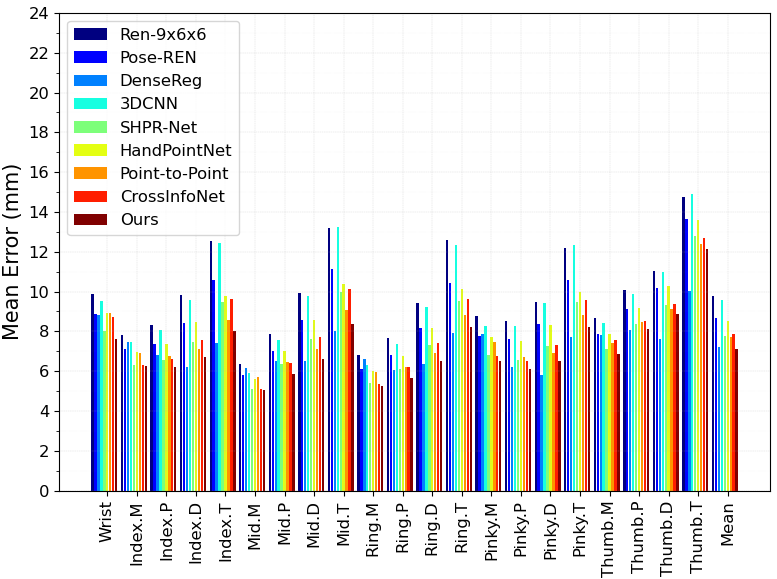} 
    \end{subfigure} 
    \caption{Comparison with the state-of-the-art methods on ICVL \cite{tang2014latent} (Left), NYU \cite{tompson2014real} (Middle), and MSRA \cite{sun2015cascaded} (Right) datasets. The per-joint mean error is used for comparison (R: root, T: tip).} 
    \label{fig:perjoint}  
\end{figure} 
\begin{figure} 
    \centering 
    \begin{subfigure}[b]{0.32\linewidth}  
        \centering 
        \includegraphics[width=\textwidth]{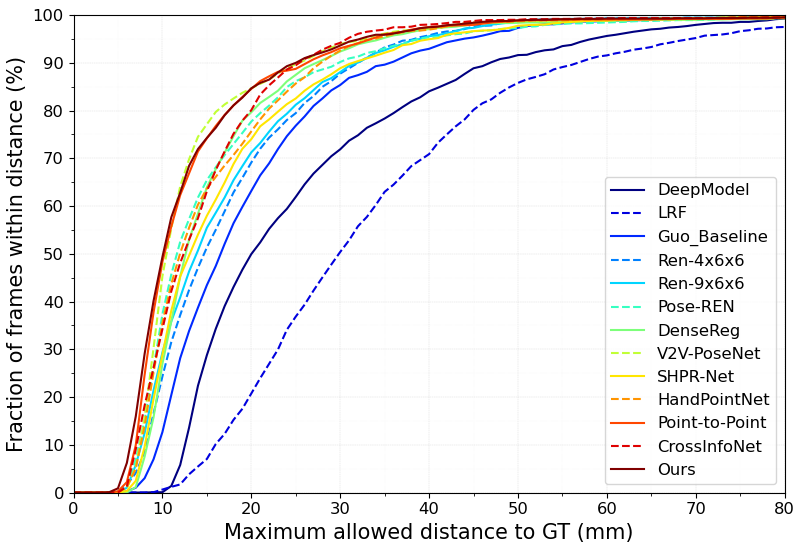} 
    \end{subfigure} 
    \begin{subfigure}[b]{0.32\linewidth} 
        \centering 
        \includegraphics[width=\textwidth]{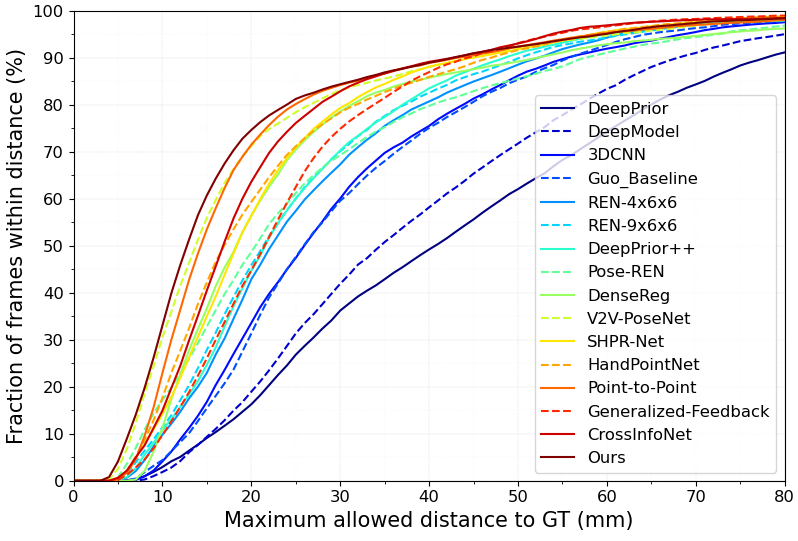} 
    \end{subfigure}
    \begin{subfigure}[b]{0.32\linewidth} 
        \centering 
        \includegraphics[width=\textwidth]{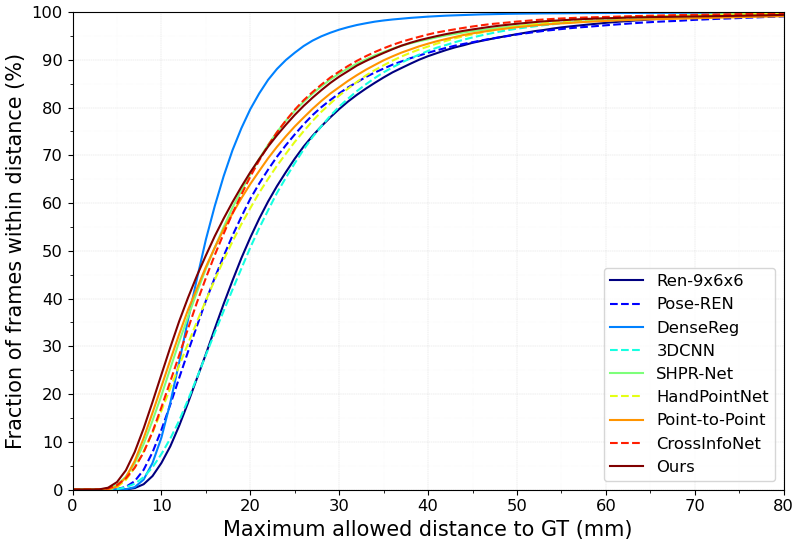} 
    \end{subfigure} 
    \caption{Comparison with the state-of-the-art methods on ICVL~\cite{tang2014latent} (Left), NYU~\cite{tompson2014real} (Middle), and MSRA~\cite{sun2015cascaded} (Right) datasets. Success rates over different error thresholds is used for comparison.} 
    \label{fig:sucrate}  
\end{figure}

\begin{table}
   \caption{Comparison with the state-of-the-art methods on ICVL~\cite{tang2014latent} (Left), NYU~\cite{tompson2014real} (Middle), and MSRA~\cite{sun2015cascaded} (Right). “Error” indicates the mean distance error in (mm)}
\begin{tabularx}{\columnwidth}{lll}
\resizebox{3cm}{!}{
    \begin{tabular}{cr}
\hline
                    Methods                    &   Error  \\
\hline
        DeepModel \cite{zhou2016model}         &        11.56 \\
      DeepPrior \cite{oberweger2015hands}      &        10.4  \\
  DeepPrior++ \cite{oberweger2017deepprior++}  &         8.1  \\
        REN-4x6x6 \cite{guo2017region}         &         7.63 \\
        REN-9x6x6 \cite{wang2018region}        &         7.31 \\
         DenseReg \cite{wan2018dense}          &         7.3  \\
         SHPR-Net \cite{chen2018shpr}          &         7.22 \\
        HandPointNet \cite{ge2018hand}         &         6.94 \\
         Pose-REN \cite{chen2020pose}          &         6.79 \\
    CrossInfoNet \cite{du2019crossinfonet}     &         6.73 \\
          NARHT \cite{huang2020hand}           &         6.47 \\
            A2J \cite{xiong2019a2j}            &         6.46 \\
       Point-to-Point \cite{ge2018point}       &         6.3  \\
        V2V-PoseNet \cite{moon2018v2v}         &         6.28 \\
          JGR-P2O \cite{fang2020jgr}           &         6.02 \\
 HandFoldingNet \cite{cheng2021handfoldingnet} &         5.95 \\
 Ours                                          &        \textbf{5.73} \\
\hline
\end{tabular}
}
&
\resizebox{3cm}{!}{
    \begin{tabular}{cr}
\hline
                       Methods                        &   Error  \\
\hline
         DeepPrior \cite{oberweger2015hands}          &        19.73 \\
            DeepModel \cite{zhou2016model}            &        17.04 \\
                3DCNN \cite{ge20173d}                 &        14.1  \\
            REN-4x6x6 \cite{guo2017region}            &        13.39 \\
           REN-9x6x6 \cite{wang2018region}            &        12.69 \\
     DeepPrior++ \cite{oberweger2017deepprior++}      &        12.24 \\
             Pose-REN \cite{chen2020pose}             &        11.81 \\
 Generalized-Feedback \cite{oberweger2019generalized} &        10.89 \\
             SHPR-Net \cite{chen2018shpr}             &        10.78 \\
            HandPointNet \cite{ge2018hand}            &        10.54 \\
             DenseReg \cite{wan2018dense}             &        10.2  \\
        CrossInfoNet \cite{du2019crossinfonet}        &        10.08 \\
              NARHT \cite{huang2020hand}              &         9.8  \\
          Point-to-Point \cite{ge2018point}           &         9.1  \\
               A2J \cite{xiong2019a2j}                &         8.61 \\
    HandFoldingNet \cite{cheng2021handfoldingnet}     &         8.58 \\
            V2V-PoseNet \cite{moon2018v2v}            &         8.42 \\
              JGR-P2O \cite{fang2020jgr}              &         8.29 \\
               Ours                                   &        \textbf{7.68} \\
\hline
\end{tabular}}
&
\resizebox{3cm}{!}{
    \begin{tabular}{cr}
\hline
                    Methods                    &   Error  \\
\hline
        REN-9x6x6 \cite{wang2018region}        &         9.79 \\
             3DCNN \cite{ge20173d}             &         9.58 \\
  DeepPrior++ \cite{oberweger2017deepprior++}  &         9.5  \\
         Pose-REN \cite{chen2020pose}          &         8.65 \\
        HandPointNet \cite{ge2018hand}         &         8.5  \\
    CrossInfoNet \cite{du2019crossinfonet}     &         7.86 \\
         SHPR-Net \cite{chen2018shpr}          &         7.76 \\
       Point-to-Point \cite{ge2018point}       &         7.7  \\
        V2V-PoseNet \cite{moon2018v2v}         &         7.59 \\
          JGR-P2O \cite{fang2020jgr}           &         7.55 \\
          NARHT \cite{huang2020hand}           &         7.55 \\
 HandFoldingNet \cite{cheng2021handfoldingnet} &         7.34 \\
         DenseReg \cite{wan2018dense}          &         7.23 \\
          Ours                                 &        \textbf{7.13} \\

\hline
\end{tabular}
}
\end{tabularx}
\label{tab:comprehensive}
\end{table}

\subsection{Comparison with the State-of-the-Art Methods}
We compare the proposed model with the state-of-the-art methods including both dense detection-based methods and regression-based methods. These methods include model-based method (DeepModel) \cite{zhou2016model}, DeepPrior \cite{oberweger2015hands}, improved DeepPrior (DeepPrior++) \cite{oberweger2017deepprior++}, region ensemble network (Ren-4x6x6 \cite{guo2017region}, Ren-9x6x6 \cite{wang2018region}), Pose-Ren \cite{chen2020pose}, Generalized-Feedback \cite{oberweger2019generalized}, dense regression network (DenseReg) \cite{wan2018dense}, A2J \cite{xiong2019a2j}, CrossInfoNet \cite{du2019crossinfonet} and JGR-P2O \cite{fang2020jgr}, 3DCNN \cite{ge20173d}, SHPR-Net \cite{chen2018shpr}, HandPointNet \cite{ge2018hand}, Point-to-Point \cite{ge2018point}, NARHT \cite{huang2020hand}, HandFoldingNet \cite{cheng2021handfoldingnet} and V2V \cite{moon2018v2v}.
Figure \ref{fig:sucrate} and Figure \ref{fig:perjoint} respectively show the success rate and per-joint mean error (mm) on the ICVL, NYU and MSRA datasets. Table~\ref{tab:comprehensive} summarizes the performance based on the mean distance error on the three datasets. The results show that the proposed method significantly outperforms the state-of-the-art methods on all of these three benchmark datasets, achieving a mean distance error of 5.73~mm, 7.68~mm and 7.13~mm on ICVL, NYU and MSRA respectively. 

\section{Conclusion}
In this paper, we proposed TriHorn-Net, a novel and powerful neural network for 3D hand pose estimation from a single depth image. It achieves improved accuracy in hand pose estimation by introducing a novel formulation to decompose the 3D hand pose estimation into the estimation of 2D joint location in the image coordinate space, and the estimation of their corresponding depth values, which is guided by an attention map resulted from the fusion of two complementary attention maps computed by two separate branches. Experimental results on three challenging benchmarks demonstrate that, despite having a simple architecture and requiring no optimization approaches at test time, the proposed network outperforms the state-of-the-art methods. We also proposed a simple data augmentation method for depth-based hand pose estimation methods and presented empirical results demonstrating its effectiveness.


\end{document}